\documentclass[11pt]{article}

\usepackage[utf8]{inputenc}
\usepackage[T1]{fontenc}
\usepackage{lmodern}
\usepackage[margin=1in]{geometry}
\usepackage{microtype}
\usepackage{setspace}
\usepackage{booktabs}
\usepackage{longtable}
\usepackage{array}
\usepackage{graphicx}
\usepackage{textcomp}
\usepackage{float}
\usepackage{amsmath,amssymb}
\usepackage{hyperref}
\usepackage[numbers,sort&compress]{natbib}

\hypersetup{
  colorlinks=true,
  linkcolor=blue,
  citecolor=blue,
  urlcolor=blue
}

\title{Large Language Models Show Metacognitive Sensitivity in Medical Reasoning}

\author{
Ahmad M. Nazzal\\
\texttt{a-nazzal@outlook.com}
}

\date{}

\begin{document}
\maketitle

\begin{abstract}
Large language models (LLMs) are increasingly evaluated and used in medicine. Clinical usefulness depends not only on answer accuracy, but also on whether confidence tracks evidence quality and uncertainty. Recent work has argued that LLMs lack essential metacognition for reliable medical reasoning, but metacognition can be operationalized in different ways, including missing-answer recognition, knowledge-gap detection, and confidence sensitivity to evidence and correctness. We developed a controlled, psychophysics-inspired clinical benchmark to test first-order diagnostic choice and second-order confidence behavior in a medical LLM. The benchmark focused on probable Alzheimer-type neurocognitive disorder (AT-NCD) versus depression-related cognitive impairment (DRCI). We generated 45 synthetic vignettes that varied evidence strength, conflicting evidence, and missing information. Each vignette was presented under three prompt variants, yielding 135 trials. In a pilot run with \texttt{gpt-4.1-nano}, all trials produced valid structured outputs. Across forced-choice trials, diagnostic accuracy was 93.5\%, mean confidence was 78.4\%, and AUROC2 was 0.876. Confidence increased with evidence distance from the diagnostic boundary, decreased in missing-information conditions, and remained higher on correct than on incorrect trials after adjustment for evidence strength and prompt format. These findings indicate partial metacognitive sensitivity rather than globally uninformative confidence. However, confidence was not uniformly reliable. Errors clustered in moderate, conflicting AT-NCD cases, where the model shifted toward DRCI and retained more confidence than empirical accuracy justified. Exploratory comparison across GPT-family models suggested that newer or nominally stronger models did not necessarily show better confidence--correctness discrimination. Thus, medical-LLM confidence should be measured directly rather than inferred from benchmark accuracy or model capability alone. This study establishes a reproducible framework for evaluating evidence sensitivity, metacognitive sensitivity, and localized calibration failure in medical LLMs.

\end{abstract}

\section{Introduction}

A core feature of good clinical reasoning is knowing when the available evidence is insufficient to support a confident decision. This capacity to monitor and evaluate one’s own judgments is commonly described as metacognition, or ``thinking about thinking'' \citep{FlemingLau2014}. Metacognition is not a decorative feature of cognition. It supports safer decisions by prompting caution, further information gathering, or deferral of judgment when uncertainty is high. As large language models are increasingly studied for medical question answering, diagnosis, triage, and decision support, it is therefore important to ask not only whether a model can produce the correct answer, but also whether it can evaluate the strength of the evidence underlying that answer. A model that reaches an incorrect conclusion while signaling uncertainty may still leave room for additional information gathering or human review. By contrast, a model that reaches an incorrect conclusion while expressing unwarranted certainty may be particularly risky in high-stakes settings such as medicine. 

Early benchmark work showed that LLMs can encode substantial clinical knowledge and perform strongly on structured medical tasks, including MultiMedQA and Med-PaLM. At the same time, those studies emphasized that clinical usefulness cannot be judged by accuracy alone and must also consider safety, factuality, bias, and reliability \citep{Singhal2023}. A central limitation of many medical-LLM evaluations is that they ask whether a model gives the correct answer, but not whether the model appropriately represents uncertainty. \citet{Hager2024} argued that examination-style medical benchmarks do not adequately test realistic clinical decision-making, including handling incomplete information, integrating evidence, and recognizing when additional information is needed. 

The uncertainty problem is now well established. \citet{Griot2025}, using MetaMedQA, showed that models lack essential metacognitive abilities for medical reasoning where the models can perform well on standard medical questions while still failing to recognize missing answers, unanswerable items, or gaps in their own knowledge. \citet{Savage2025} found that directly elicited verbal confidence can overestimate reliability in medical diagnosis and treatment tasks. \citet{Gu2024} similarly reported that LLMs struggle to produce reliable explicit medical probability estimates. Together, these studies suggest that first-order task performance and second-order self-evaluation may come apart in clinically relevant ways.

This problem connects naturally to the metacognition literature. In cognitive science, confidence is treated as a second-order estimate of the probability that a choice is correct. What matters is not merely whether confidence is high or low, but whether it tracks evidence quality and correctness. \citet{FlemingLau2014} argue that metacognitive sensitivity should be assessed through the relation between confidence and performance rather than by mean confidence alone. In parallel, calibration research in machine learning asks whether stated confidence corresponds to observed correctness probability \citep{FlemingLau2014,Guo2017}. For signal-detection-theoretic work, \citet{ManiscalcoLau2012} further distinguish metacognitive sensitivity from metacognitive efficiency, the latter indexing how much of the information available to the primary decision process is reflected in confidence reports.

Most medical-LLM studies still evaluate uncertainty in heterogeneous benchmark settings. That makes it hard to determine whether confidence reflects evidence strength, prompt framing, prior bias, or task-specific artifacts. Psychophysics offers a more controlled alternative. In psychophysical paradigms, evidence strength is manipulated systematically, and observer behavior is analyzed in terms of threshold, slope, bias, and uncertainty. \citet{WichmannHill2001} formalized this approach for psychometric-function estimation. The same general logic can be applied to second-order evaluation: if confidence is meaningful, it should vary with evidence and correctness in predictable ways \citep{WichmannHill2001,FlemingLau2014}.

The present study applies that logic to medical LLMs. Rather than asking whether a model can answer broad medical questions, we introduce a controlled clinical-evidence benchmark in which synthetic vignettes vary in the amount and quality of evidence favoring one of two competing diagnostic interpretations. We focus on a single clinically relevant contrast: probable Alzheimer-type neurocognitive disorder (AT-NCD) versus depression-related cognitive impairment (DRCI). This differential was chosen because it is medically meaningful, cognitively rich, and well suited to systematic manipulation of ambiguity, conflicting evidence, and missing information. Clinically, it captures a familiar diagnostic tension: progressive neurocognitive decline may overlap with or be obscured by depressive symptoms. \citet{McKhann2011} provide the general clinical framework for Alzheimer-type dementia syndromes, while work on late-life depression shows that depression can be accompanied by substantial cognitive impairment, especially in executive function, processing speed, and memory complaints \citep{McKhann2011,Koenig2014}.

Our aim is not to determine whether LLMs possess ``true'' metacognition in any broad philosophical sense. The narrower goal is behavioral: to test whether explicit confidence in a diagnostic choice tracks clinical evidence strength, information quality, and correctness. Specifically, we ask three questions. First, do model choices vary systematically with graded evidence favoring AT-NCD or DRCI? Second, does reported confidence increase as cases move away from the diagnostic boundary and decrease when information is missing or conflicting? Third, does confidence remain informative about correctness after accounting for evidence strength itself?

By combining controlled vignette generation, structured confidence elicitation, and condition-level calibration analyses, this study aims to establish a reproducible framework for probing diagnostic confidence in medical LLMs. The central premise is simple: if confidence is to be clinically meaningful, it should not merely accompany a diagnosis, but should vary appropriately with the quality and sufficiency of the underlying evidence.

\begin{figure}[H]
    \centering
   \includegraphics[width=\linewidth]{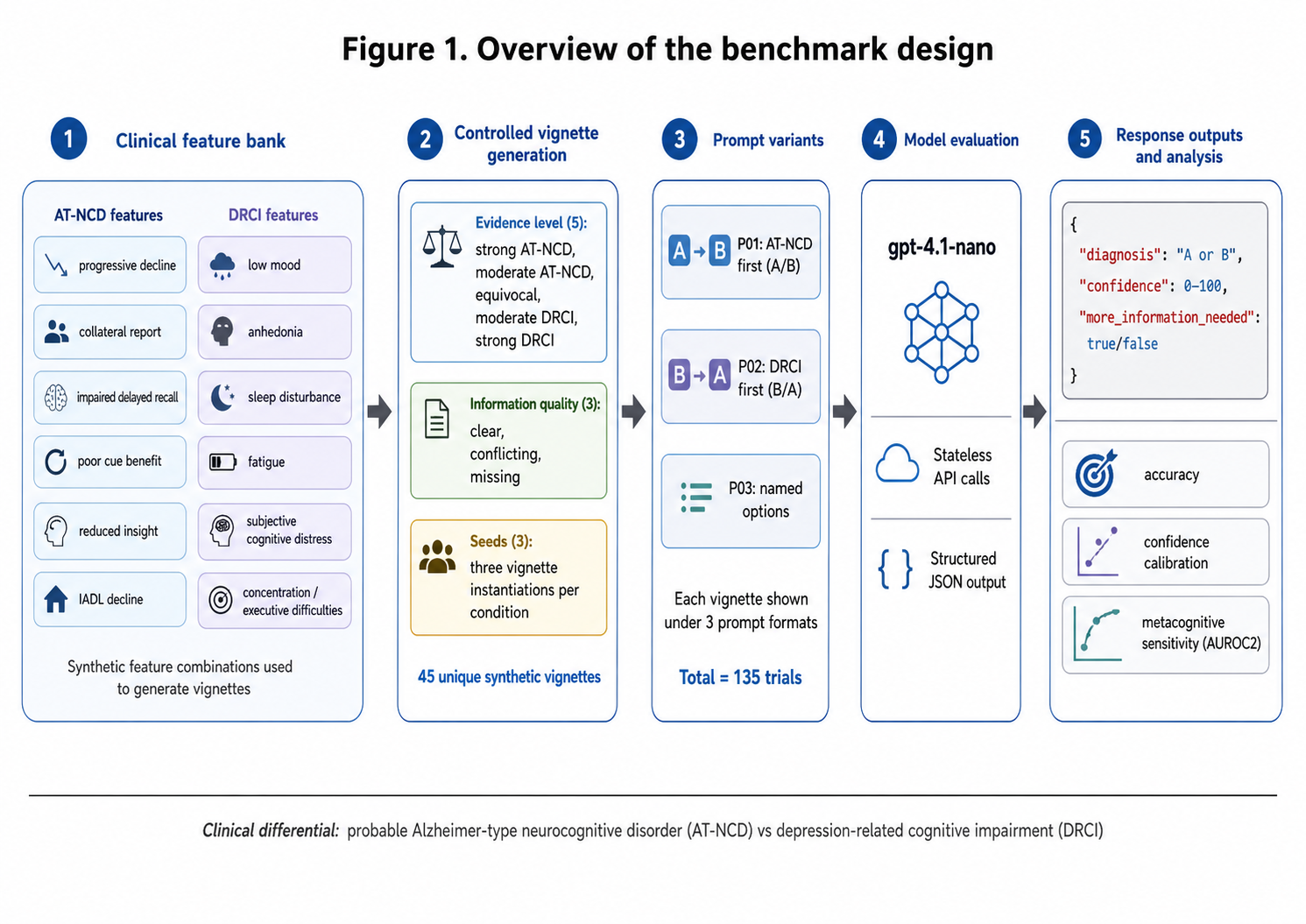}
    \caption{Overview of the benchmark design.}
    \label{fig:overview}
\end{figure}

\section{Methods}

\subsection{Study design}

We designed a controlled, psychophysics-inspired benchmark to test whether a medical LLM's diagnostic choices and reported confidence vary systematically with the strength and quality of clinical evidence. Rather than evaluating broad medical knowledge across heterogeneous questions, we treated the model as a behavioral observer and manipulated the amount of evidence favoring one of two competing diagnostic interpretations. This design follows the logic of psychophysics, in which observer performance is analyzed as a function of experimentally controlled evidence strength \citep{WichmannHill2001}. The benchmark focused on one clinical differential: probable Alzheimer-type neurocognitive disorder (AT-NCD) versus depression-related cognitive impairment (DRCI). This contrast was selected because it is clinically plausible and suitable for controlled manipulation of ambiguity, conflicting evidence, and missing information.

\subsection{Diagnostic framing}

The label \emph{probable Alzheimer-type neurocognitive disorder} was used as an operational syndrome-level label for synthetic vignette generation. It does not imply biomarker-confirmed Alzheimer's disease. This choice preserves the central contrast between progressive Alzheimer-type decline and depression-related cognitive impairment while remaining clinically cautious. The AT-NCD side of the benchmark was informed by the NIA-AA framework for dementia due to Alzheimer's disease, which emphasizes insidious onset, gradual progression, decline from prior functioning, objective impairment, and interference with daily life \citep{McKhann2011}. The DRCI side was informed by literature on late-life depression and cognition, especially work showing that depression in older adults can produce clinically relevant cognitive impairment \citep{Koenig2014}.

\subsection{Benchmark construction}

The benchmark contained 45 unique synthetic clinical vignettes, each paired with 3 prompt variants, yielding 135 total trials. Vignettes were generated from a predefined feature bank containing clinically plausible cues favoring either AT-NCD or DRCI, along with neutral features, conflict cues, and missing-information cues. Internal feature weights were used only for vignette construction and analysis metadata; they were not shown to the model. The 45 vignettes were organized as a factorial set spanning (i) five evidence levels: strong AT-NCD, moderate AT-NCD, equivocal, moderate DRCI, and strong DRCI; (ii) three information-quality conditions: clear, conflicting, and missing; and (iii) three vignette seeds per condition.

\subsection{Feature bank and evidence levels}

Features supporting AT-NCD included gradual progression over months to years, collateral reports of decline, impaired delayed recall, poor cue benefit, reduced insight, and functional decline in instrumental activities. Features supporting DRCI included low mood, anhedonia, sleep disturbance, fatigue, subjective distress about cognition, variable effort, concentration difficulties, and patterns more consistent with slowed processing or executive dysfunction than stable amnestic storage failure. This feature-bank approach was intended to produce controlled clinical evidence perturbations rather than encode a clinical guideline. For each vignette, an internal evidence score was calculated as the sum of weighted features favoring AT-NCD versus DRCI. These scores were then grouped into five ordered evidence levels: strong AT-NCD, moderate AT-NCD, equivocal, moderate DRCI, and strong DRCI.

\subsection{Information-quality conditions}

Each evidence level was crossed with three information-quality conditions. In the \emph{clear} condition, vignettes contained internally coherent evidence favoring one interpretation. In the \emph{conflicting} condition, vignettes included clinically plausible cues supporting both AT-NCD and DRCI. In the \emph{missing} condition, one or more clinically informative elements were omitted, such as collateral history, longitudinal information, cognitive testing, or structured mood assessment. This manipulation was motivated by the fact that clinical uncertainty arises not only from weak evidence, but also from competing and incomplete evidence. It also follows prior work arguing that realistic medical-LLM evaluation should go beyond closed-book multiple-choice recall and examine performance under incomplete or ambiguous clinical information \citep{Hager2024}.

\subsection{Prompt design}

Each vignette was presented with a standardized prompt asking the model to choose between the two diagnostic interpretations, provide a numerical confidence estimate, and indicate whether more information was needed. Three prompt variants were used to test prompt-format effects: P01, AT-NCD listed first with lettered options; P02, DRCI listed first with lettered options; and P03, named options with AT-NCD first. The model was instructed to return a JSON object containing \texttt{diagnosis} (\texttt{A} or \texttt{B}), \texttt{confidence} (0--100), and \texttt{more\_information\_needed} (\texttt{true} or \texttt{false}). Confidence was explicitly defined as the model's estimated probability that its chosen interpretation was correct based only on the information provided.

\subsection{Model and inference settings}

The primary pilot reported here used \texttt{gpt-4.1-nano}, accessed through the OpenAI Responses API. All calls were stateless. No conversation history was retained across trials. The prompt was submitted using Structured Outputs with a strict JSON schema, so that the model response was constrained to the three required fields. Raw outputs and parsed outputs were saved trial by trial.

\subsection{Outcome measures}

We defined first-order and second-order outcomes separately. First-order outcomes were forced-choice diagnostic accuracy for non-equivocal cases and evidence-sensitive diagnostic choice behavior across the clinical evidence gradient. Second-order outcomes were metacognitive bias, metacognitive sensitivity, and exploratory metacognitive efficiency. Metacognitive bias was operationalized as the relation between mean confidence and empirical accuracy. Metacognitive sensitivity was operationalized primarily as AUROC2 and secondarily as the relation between confidence and correctness in regression models. Metacognitive efficiency was examined exploratorily using SDT-derived quantities, including $d^\prime$, meta-$d^\prime$, and their ratio, meta-$d^\prime$/$d^\prime$, following the framework described by \citet{FlemingLau2014} and \citet{ManiscalcoLau2012}. We also evaluated the practical uncertainty judgment conveyed by the \texttt{more\_information\_needed} variable.

\subsection{Calibration and confidence metrics}

Global calibration was assessed using mean confidence, Brier score, expected calibration error (ECE), and reliability diagrams. Because the benchmark elicited explicit probability-like confidence judgments, these measures were used to quantify the relation between stated confidence and empirical correctness. Metacognitive sensitivity was quantified primarily using AUROC2, which measures how well confidence discriminates correct from incorrect responses independent of raw confidence magnitude. In addition, confidence was modeled as a function of correctness after adjustment for evidence strength, information quality, and prompt format. Exploratory SDT-based analyses were also performed to characterize first-order sensitivity and metacognitive efficiency. Type-1 sensitivity was summarized using $d^\prime$. Metacognitive efficiency was examined using meta-$d^\prime$ and the ratio meta-$d^\prime$/$d^\prime$. Because the pilot dataset was small, several cells were near ceiling, and confidence values occupied a restricted range, these SDT-based efficiency analyses were treated as exploratory rather than primary.

\subsection{Statistical analysis}

All analyses were performed in Python. We first computed condition-level summary tables for parsed response rate, mean confidence, forced-choice accuracy, the proportion of \texttt{more\_information\_needed = true}, and information-sufficiency accuracy. Diagnostic choice was summarized primarily through condition-level results and the accuracy-by-evidence curve. As a robustness check, we also fitted a penalized logistic model in which AT-NCD choice was predicted by evidence score, information-quality condition, and prompt variant. Because several conditions were solved at or near ceiling, this model was treated as supportive rather than central and is reported in the Supplement. Confidence was modeled using ordinary least squares regression. In the primary confidence model, predictors were evidence distance from the diagnostic boundary, information-quality condition, and prompt variant. To evaluate confidence--correctness coupling, we fitted a second model restricted to forced-choice trials, with predictors for correctness, evidence distance, information quality, and prompt variant. Because each vignette appeared in three prompt variants, these analyses were repeated with cluster-robust standard errors clustered by vignette ID.

\subsection{Equivocal cases}

Equivocal vignettes were designed to be underdetermined. For these cases, forced-choice accuracy was treated as not applicable. The principal outcomes were whether the model indicated that additional information was needed and whether confidence was reduced appropriately.

\subsection{Reproducibility and ethics}

The benchmark was designed as a reproducible pilot. The feature bank, vignette generator, prompt templates, trial manifest, raw outputs, parsed outputs, and analysis notebooks were versioned and frozen after the run. No patient data were used. All cases were synthetic and intended solely for model evaluation. The benchmark is not a clinical tool and does not provide medical advice.

\begin{table}[H]
\centering
\caption{Benchmark design.}
\label{tab:design}
\begin{tabular}{p{0.35\linewidth} p{0.55\linewidth}}
\toprule
Component & Specification \\
\midrule
Clinical differential & Probable Alzheimer-type neurocognitive disorder (AT-NCD) vs depression-related cognitive impairment (DRCI) \\
Benchmark type & Controlled synthetic clinical-evidence benchmark \\
Unique vignettes & 45 \\
Prompt variants per vignette & 3 \\
Total trials per model & 135 \\
Evidence levels & Strong AT-NCD, Moderate AT-NCD, Equivocal, Moderate DRCI, Strong DRCI \\
Information-quality conditions & Clear, Conflicting, Missing \\
Equivocal cases & Treated as underdetermined; forced-choice accuracy not applicable \\
Model outputs required & Diagnosis, confidence (0--100), more-information-needed judgment \\
Main first-order outcome & Forced-choice diagnostic accuracy \\
Main second-order outcomes & Confidence calibration, AUROC2, information-sufficiency accuracy \\
\bottomrule
\end{tabular}
\end{table}

\section{Results}

\subsection{Benchmark integrity and output validity}

The benchmark ran on the full 135-trial manifest. The model produced valid structured outputs on all trials, with no malformed responses or parsing failures. Thus, the JSON-constrained inference pipeline was stable. Of the 135 trials, 108 were forced-choice trials in which diagnostic accuracy was applicable. The remaining 27 corresponded to equivocal or explicitly underdetermined cases, for which the primary uncertainty outcome was whether the model indicated that more information was needed.

\subsection{Overall performance of the primary pilot model gpt-4.1-nano}

Across forced-choice trials, \texttt{gpt-4.1-nano} achieved 93.5\% accuracy. Mean confidence was 78.4\%. Global calibration was imperfect but not grossly poor (Brier score = 0.077; ECE = 0.151). Confidence discriminated correct from incorrect responses well at the aggregate level (AUROC2 = 0.876). Accuracy on the information-sufficiency judgment was 83.7\%.

These aggregate metrics indicate that the model's confidence outputs were not arbitrary. However, aggregate performance did not capture the full pattern.

\begin{table}[H]
\centering
\caption{Overall performance for \texttt{gpt-4.1-nano}.}
\label{tab:overall}
\begin{tabular}{lr}
\toprule
Metric & Value \\
\midrule
Total trials & 135 \\
Forced-choice trials & 108 \\
Malformed / parse error rate & 0.000 \\
Forced-choice accuracy & 0.935 \\
Mean confidence & 0.784 \\
Brier score & 0.077 \\
ECE & 0.151 \\
AUROC2 & 0.876 \\
More-information-needed accuracy & 0.837 \\
\bottomrule
\end{tabular}
\end{table}

\begin{figure}[H]
    \centering
    \includegraphics[width=0.6\linewidth]{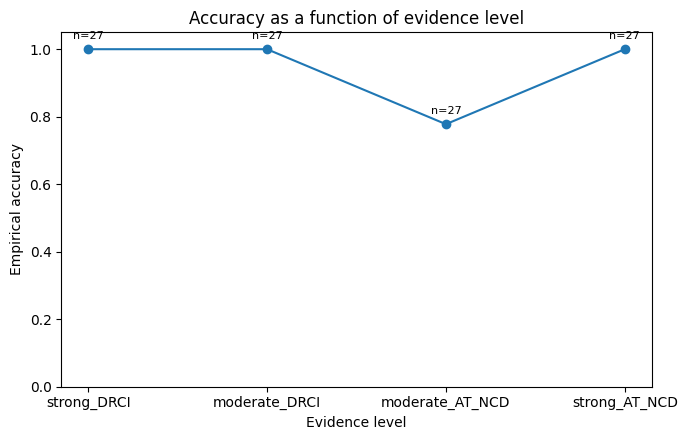}
    \caption{Accuracy as a function of evidence level. Empirical accuracy is shown across forced-choice trials only. The equivocal condition is included on the x-axis for completeness but is not assigned an accuracy value because it was intentionally designed as underdetermined.}
    \label{fig:accuracy_evidence}
\end{figure}

\subsection{Condition-level performance revealed a localized failure zone}

The benchmark was not uniformly difficult. Nearly all strong-evidence conditions were solved at or near ceiling. Moderate DRCI conditions were also generally classified correctly. Errors were concentrated in a narrow region of the diagnostic space.

The clearest failure occurred in the moderate AT-NCD plus conflicting-information condition. In this cell, forced-choice accuracy dropped to 44.4\%, while mean confidence remained 72.8\%. This yielded an overconfidence gap of 28.3 percentage points, making it the most pronounced local calibration failure in the pilot. By contrast, the corresponding moderate DRCI conflicting condition was classified correctly on all trials.

Equivocal cases also showed an asymmetry. When the evidence was underdetermined, the model tended to default toward DRCI rather than AT-NCD. Thus, ambiguity did not collapse into random guessing. Instead, it shifted the decision boundary toward a specific diagnostic default.

\begin{table}[H]
\centering
\caption{Condition-level summary for \texttt{gpt-4.1-nano}.}
\label{tab:conditions}
\begin{tabular}{llrrrrr}
\toprule
Evidence level & Info quality & $n$ & Mean conf. & Accuracy & Overconf. & More-info acc. \\
\midrule
Equivocal & Clear & 9 & 0.700 & --- & --- & 1.000 \\
Equivocal & Conflicting & 9 & 0.706 & --- & --- & 1.000 \\
Equivocal & Missing & 9 & 0.644 & --- & --- & 1.000 \\
Moderate AT-NCD & Clear & 9 & 0.850 & 1.000 & -0.150 & 1.000 \\
Moderate AT-NCD & Conflicting & 9 & 0.728 & 0.444 & 0.283 & 0.778 \\
Moderate AT-NCD & Missing & 9 & 0.689 & 0.778 & -0.089 & 1.000 \\
Moderate DRCI & Clear & 9 & 0.806 & 1.000 & -0.194 & 0.333 \\
Moderate DRCI & Conflicting & 9 & 0.756 & 1.000 & -0.244 & 0.778 \\
Moderate DRCI & Missing & 9 & 0.711 & 1.000 & -0.289 & 1.000 \\
Strong AT-NCD & Clear & 9 & 0.850 & 1.000 & -0.150 & 1.000 \\
Strong AT-NCD & Conflicting & 9 & 0.811 & 1.000 & -0.189 & 0.333 \\
Strong AT-NCD & Missing & 9 & 0.778 & 1.000 & -0.222 & 1.000 \\
Strong DRCI & Clear & 9 & 0.850 & 1.000 & -0.150 & 1.000 \\
Strong DRCI & Conflicting & 9 & 0.828 & 1.000 & -0.172 & 0.333 \\
Strong DRCI & Missing & 9 & 0.750 & 1.000 & -0.250 & 1.000 \\
\bottomrule
\end{tabular}
\end{table}

\subsection{Confidence was evidence-sensitive and showed partial metacognitive sensitivity}

Confidence was not globally flat. Descriptively, confidence was highest in strong clear-evidence conditions, lower in moderate conditions, lower again in missing-information conditions, and lowest in equivocal missing-information cases.

\begin{figure}[H]
    \centering
    \includegraphics[width=0.6\linewidth]{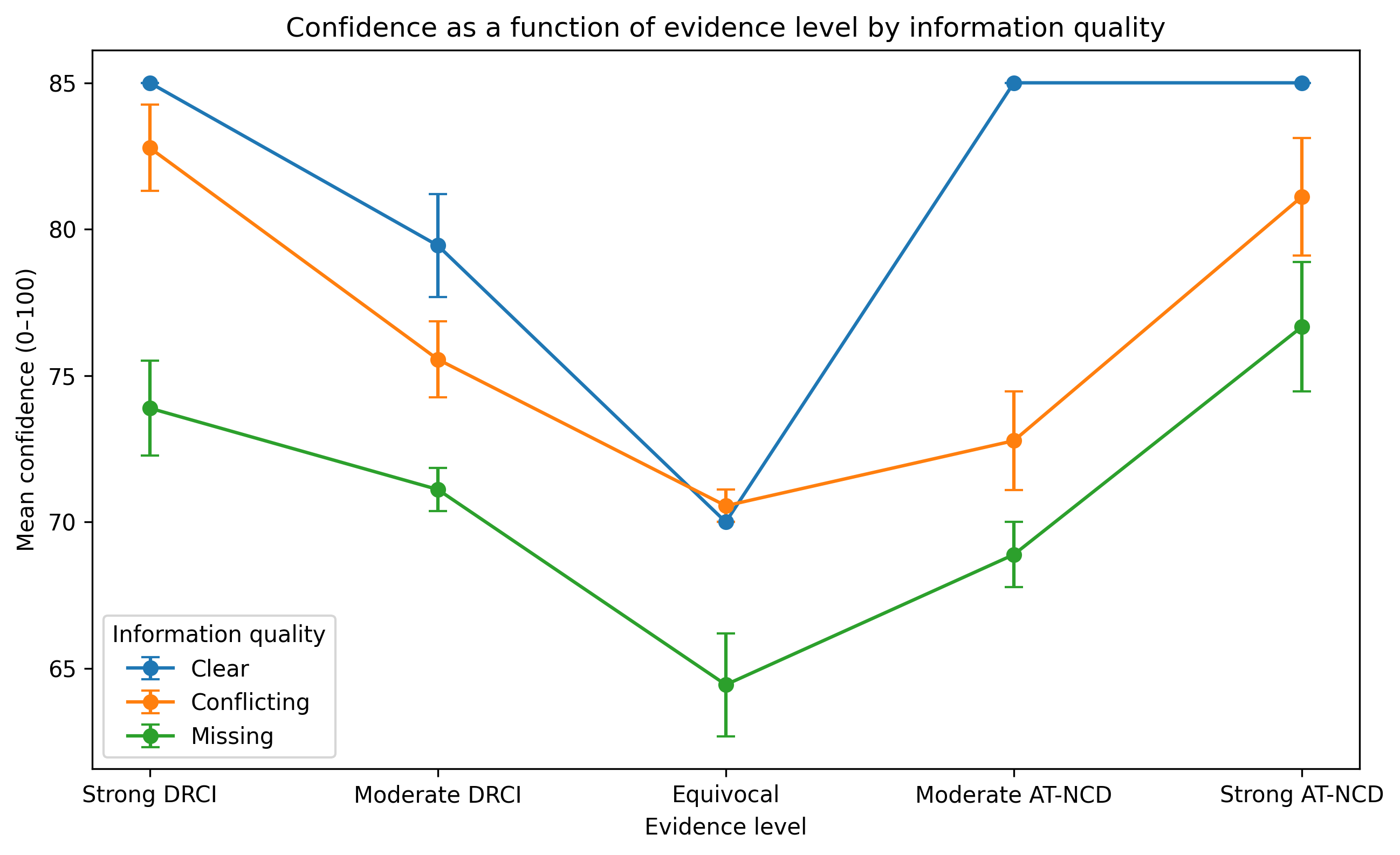}
    \caption{Confidence as a function of evidence level, stratified by information quality. Mean confidence is shown across all parsed trials. Error bars indicate the standard error of the mean.}
    \label{fig:confidence_evidence}
\end{figure}

A linear model confirmed this pattern. Confidence increased significantly with distance from the diagnostic boundary (coefficient = 1.08, $p < 0.001$). Missing-information conditions reduced confidence by approximately 4.83 points ($p < 0.001$). In contrast, conflicting information did not significantly reduce confidence ($p = 0.94$). Thus, the model responded appropriately to absent information, but not to internally competing evidence.

Prompt format also influenced confidence. Relative to the reference prompt, the named-options format reduced confidence by approximately 3.22 points ($p = 0.002$), indicating that confidence was partly elicitation-dependent.

To test whether confidence tracked correctness rather than only evidence magnitude, we fitted a cluster-robust regression restricted to forced-choice trials. Confidence remained significantly higher on correct than on incorrect trials even after adjustment for evidence strength, information quality, and prompt format (coefficient for correctness = +5.15, $p = 0.001$). Evidence distance remained a significant positive predictor of confidence, and missing-information conditions remained associated with lower confidence.

Framed in metacognitive terms, the model showed evidence of second-order sensitivity. Confidence discriminated correct from incorrect responses well at the aggregate level (AUROC2 = 0.876), and confidence remained significantly associated with correctness after adjustment for first-order difficulty. This indicates that confidence was not merely a by-product of evidence magnitude. Instead, confidence retained some independent relationship to correctness.

\begin{figure}[H]
    \centering
    \includegraphics[width=0.5\linewidth]{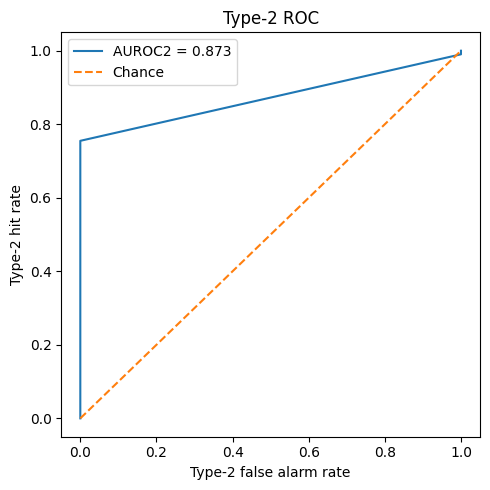}
    \caption{Type-2 ROC curve for confidence--correctness discrimination. The curve summarizes how well confidence discriminates correct from incorrect forced-choice responses. Because confidence values were discrete and incorrect trials were sparse, the ROC curve is coarse rather than smooth. AUROC2 was 0.873, indicating substantial confidence--correctness discrimination.}
    \label{fig:type2roc}
\end{figure}

\begin{figure}[H]
    \centering
    \includegraphics[width=0.7\linewidth]{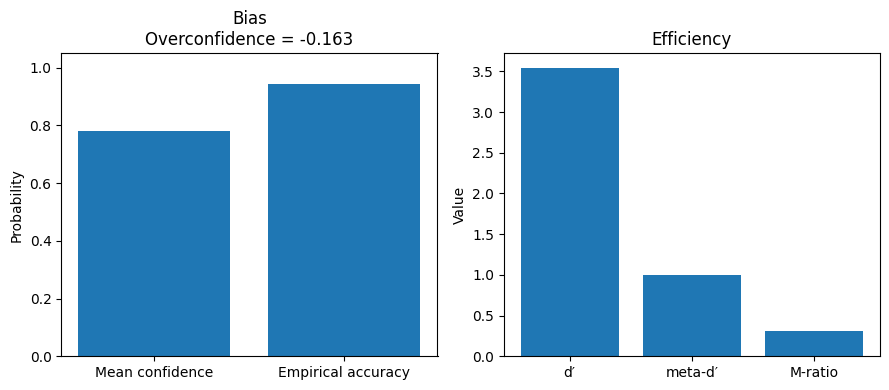}
    \caption{Summary of metacognitive bias, sensitivity, and exploratory efficiency. Figure shows mean confidence versus empirical accuracy, and exploratory SDT-based efficiency indices.}
    \label{fig:metacog_summary}
\end{figure}

\begin{table}[H]
\centering
\caption{Primary regression summary for \texttt{gpt-4.1-nano}.}
\label{tab:regression}
\begin{tabular}{lrrr}
\toprule
Predictor & Coefficient & SE & $p$ \\
\midrule
\multicolumn{4}{l}{\textit{Primary confidence model}} \\
Intercept & 72.067 & 1.445 & $<0.001$ \\
Conflicting information & -0.084 & 1.128 & 0.941 \\
Missing information & -4.835 & 1.144 & $<0.001$ \\
DRCI-first prompt (P02) & -1.667 & 1.030 & 0.108 \\
Named-options prompt (P03) & -3.222 & 1.030 & 0.002 \\
Evidence distance & 1.082 & 0.111 & $<0.001$ \\
\midrule
\multicolumn{4}{l}{\textit{Cluster-robust model restricted to forced-choice trials}} \\
Intercept & 74.151 & 2.146 & $<0.001$ \\
Conflicting information & -2.080 & 1.921 & 0.279 \\
Missing information & -7.638 & 1.895 & $<0.001$ \\
DRCI-first prompt (P02) & -1.381 & 0.890 & 0.121 \\
Named-options prompt (P03) & -3.754 & 0.878 & $<0.001$ \\
Correct response & 5.150 & 1.531 & 0.001 \\
Evidence distance & 0.536 & 0.192 & 0.005 \\
\bottomrule
\end{tabular}
\end{table}

\subsection{The main calibration problem was local rather than global}

The main second-order problem was local rather than global. At the aggregate level, the model was mildly underconfident rather than overconfident. Most forced-choice conditions showed empirical accuracy of 100\%, whereas mean confidence was typically between 75\% and 85\%. Thus, the model was generally conservative overall.

The major exception was the moderate AT-NCD plus conflicting-information condition. In this failure zone, empirical accuracy was 44.4\%, but mean confidence remained 72.8\%. Within that cell, incorrect trials had a mean confidence of 70.0\%, whereas correct trials had a mean confidence of 76.25\%. Confidence therefore retained some discriminative value within the problematic condition, but not enough to prevent a large local calibration breakdown.

The main conclusion from the results is therefore not that medical-LLM confidence is globally useless. Rather, the results suggest that confidence is partially calibrated overall but locally fragile, with a clinically meaningful breakdown at a specific ambiguity boundary where Alzheimer-type and depressive features compete.

\begin{table}[H]
\centering
\caption{Focused analysis of the principal failure condition.}
\label{tab:failure_zone}
\small
\setlength{\tabcolsep}{4pt}
\begin{tabular}{lrrrrrr}
\toprule
Condition & $n_{\text{corr}}$ & $n_{\text{incorr}}$ & Conf.\ corr. & Conf.\ incorr. & Acc. & Overconf. \\
\midrule
Moderate AT-NCD + conflicting & 4 & 5 & 0.763 & 0.700 & 0.444 & 0.283 \\
\bottomrule
\end{tabular}
\end{table}

\subsection{Exploratory cross-model comparison}

As an exploratory comparison, we repeated the same 135-trial benchmark using four additional GPT-family models: \texttt{gpt-4.1-mini}, \texttt{gpt-5-nano}, \texttt{gpt-5}, and \texttt{gpt-5.5}. Together with the primary \texttt{gpt-4.1-nano} run, all models produced valid parsed outputs for the full manifest. Each run yielded 108 forced-choice trials. We compared metacognitive sensitivity using AUROC2, defined as the ability of confidence to discriminate correct from incorrect forced-choice responses.

AUROC2 differed across models. Among models for which AUROC2 was estimable, \texttt{gpt-5} showed the highest confidence--correctness discrimination (AUROC2 = 0.919), followed by \texttt{gpt-4.1-nano} (AUROC2 = 0.873), \texttt{gpt-4.1-mini} (AUROC2 = 0.786), and \texttt{gpt-5-nano} (AUROC2 = 0.644). The corresponding forced-choice accuracies were 0.944 for \texttt{gpt-5}, 0.944 for \texttt{gpt-4.1-nano}, 0.963 for \texttt{gpt-4.1-mini}, and 0.713 for \texttt{gpt-5-nano}. Thus, in this pilot benchmark, confidence--correctness discrimination did not increase monotonically with nominal model capability or model family. In particular, \texttt{gpt-4.1-mini} achieved higher diagnostic accuracy than \texttt{gpt-4.1-nano}, but lower AUROC2.

We also evaluated \texttt{gpt-5.5}. This model achieved perfect forced-choice accuracy (1.000), with mean confidence of 0.786 and Brier score of 0.063. Because it made no forced-choice errors, AUROC2 was undefined: there were no incorrect trials against which confidence could be discriminated. However, confidence still varied across evidence levels and information-quality conditions (Supplementary Figure~1), showing that confidence modulation by evidence and confidence--correctness discrimination are related but distinct quantities. This ceiling effect suggests that stronger models will require a harder benchmark around the diagnostic ambiguity boundary.

\begin{figure}[H]
    \centering
    \includegraphics[width=0.7\linewidth]{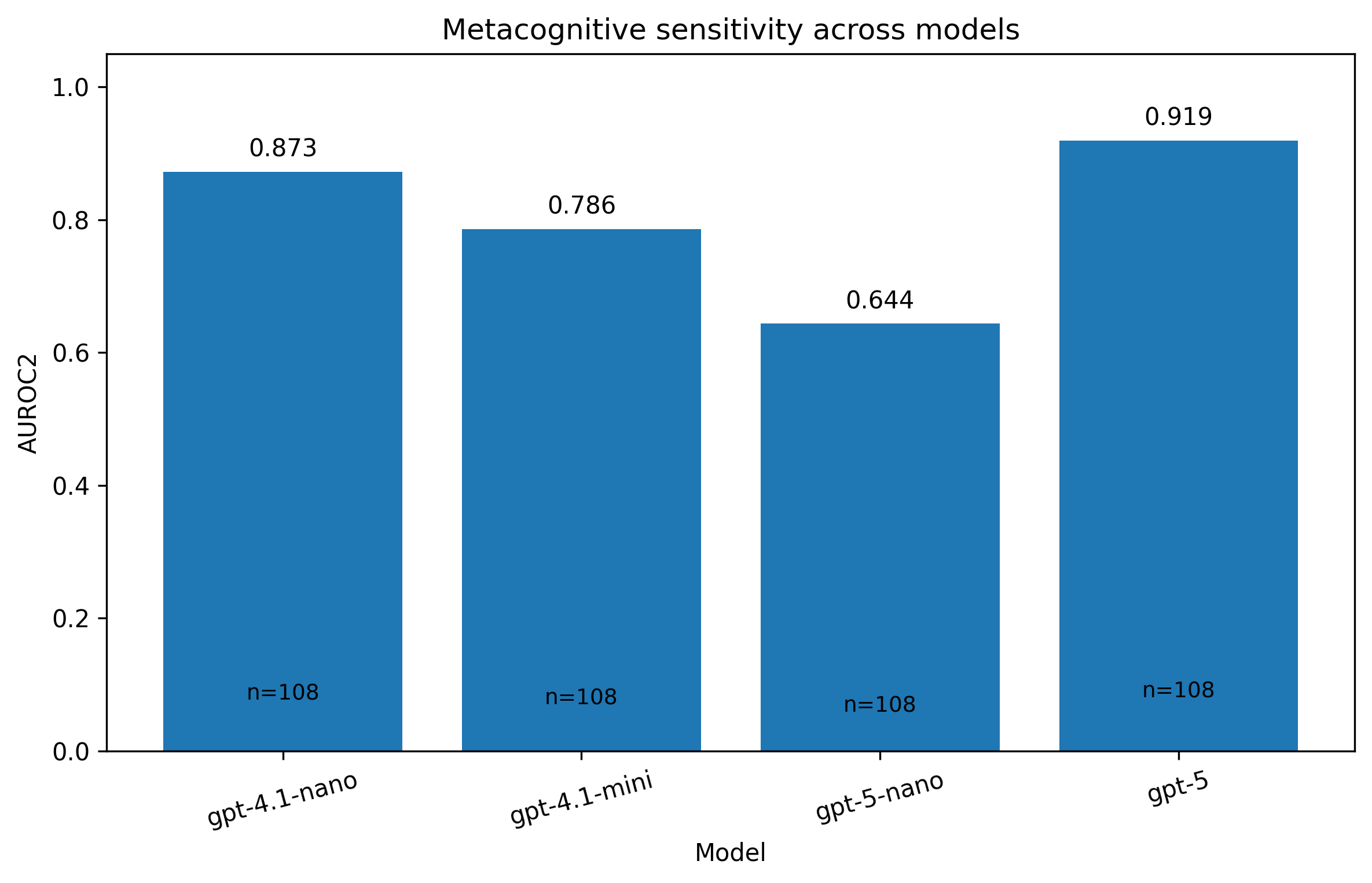}
    \caption{Exploratory model comparison of metacognitive sensitivity, quantified using AUROC2. AUROC2 measures how well confidence discriminates correct from incorrect forced-choice responses. Higher values indicate better confidence--correctness discrimination. All models were evaluated on 108 forced-choice parsed trials.}
    \label{fig:auroc2_bar_models}
\end{figure}

\section{Discussion}

This pilot study introduces a controlled clinical-evidence paradigm for probing diagnostic choice, uncertainty behavior, and confidence calibration in a medical LLM. Three main findings emerged. First, diagnostic choice was strongly evidence-sensitive. As AT-NCD-consistent evidence increased, the probability of choosing AT-NCD increased correspondingly. Second, confidence was not arbitrary or globally flat. Confidence tracked evidence strength, decreased in missing-information conditions, and remained higher on correct than on incorrect trials after adjustment for evidence distance and prompt format. Third, these globally sensible second-order properties coexisted with a localized failure mode: in moderate, conflicting AT-NCD cases, the model tended to shift toward DRCI and retained more confidence than empirical accuracy justified.

The broadest implication is methodological. Current medical-LLM evaluation often relies on heterogeneous exam-style items or broad benchmark aggregates, which can obscure the structure of diagnostic uncertainty. Prior work has already shown that strong benchmark performance does not ensure realistic clinical reasoning, especially when tasks require incomplete information handling and explicit recognition of uncertainty \citep{Hager2024}. The present benchmark complements that literature by using a controlled evidence gradient, making it possible to separate first-order diagnostic sensitivity from second-order confidence behavior more cleanly than broad accuracy metrics alone.

 Moreover, the claim --- that medical-LLM confidence is globally uninformative --- is not supported by these data. Instead, confidence retained measurable metacognitive value: it increased with evidence distance, decreased when key information was missing, and remained higher on correct than on incorrect trials after adjustment. In the framework proposed by \citet{FlemingLau2014}, this is consistent with partial second-order sensitivity rather than pure confidence noise.

The distinction between metacognitive bias, sensitivity, and efficiency is important here. The model was globally underconfident, indicating negative metacognitive bias at the aggregate level. It nevertheless showed substantial metacognitive sensitivity, because confidence discriminated correct from incorrect responses well. Exploratory efficiency analysis suggested that this second-order signal was weaker than would be expected from first-order task performance alone. The resulting picture is therefore not one of absent metacognition, but of partial and uneven metacognitive access.

The failure pattern is clinically interpretable. Errors were concentrated in Alzheimer-type cases containing competing depressive features, and equivocal cases showed a DRCI-leaning default. One plausible interpretation is that depressive-context cues were overweighted relative to moderate but progressive neurocognitive evidence. Another is that DRCI functioned as a lower-commitment or more reversible default under uncertainty. The present design cannot distinguish those explanations mechanistically, but it does show that the boundary between progressive neurocognitive decline and depression-related cognitive impairment is exactly where confidence calibration and decision bias become most informative.

The study has several limitations. First, this was a synthetic vignette benchmark centered on one diagnostic contrast. The task was deliberately narrow, and the results should not be generalized to clinical reasoning as a whole. Second, although the benchmark included controlled manipulations of evidence strength, conflict, and missing information, the vignette bank was still small, and several conditions were solved at or near ceiling. Third, the confidence measure was explicit and prompt-elicited. Prior work suggests that direct verbalized confidence can differ from other uncertainty proxies, including sample-consistency methods or internal likelihood-based estimates \citep{Savage2025,Gu2024}. Future work could compare both. Fourth, the \texttt{more\_information\_needed} variable worked reasonably well as a practical uncertainty readout, but conceptually it mixes underdetermination, caution, and response thresholding. A refined future version should distinguish ``case underdetermined'' from ``additional information would be useful.''

This preprint should be read as a controlled proof of concept showing that medical-LLM confidence can be studied with psychophysics-inspired methods and that such methods can reveal patterns not visible in aggregate benchmark scores.

In this study, we use the language of metacognition in a behavioral and operational sense, not in the full human cognitive sense. In humans, metacognition refers to the capacity to monitor and evaluate one’s own mental states, judgments, and decision processes. LLMs can be prompted to comment on their own answers, estimate confidence, or revise a response, and this can make their behavior appear superficially reflective. However, such outputs should not be taken as evidence that the model is thinking about its own thinking in the human sense. The mechanisms underlying LLM behavior are statistical, and the relationship between language generation, evidence use, uncertainty estimation, and any internal model state remains incompletely understood.

The claim made here is therefore narrower and more practical. We ask whether a model’s expressed confidence varies systematically with the evidence provided in the prompt and with the correctness of its own output. In this operational sense, psychophysics- and metacognition-inspired measures can be useful even if the model does not possess human-like reflective awareness. They provide a structured behavioral way to test whether the model can use available evidence to modulate certainty, recognize underdetermined cases, and signal when its answer may be unreliable. This kind of framework may be especially valuable in medicine, education, and the social sciences, where trust, uncertainty, and decision support are central concerns and where purely technical metrics may not fully capture the practical risks of model use.

This framing also opens a path beyond language-only tasks. The present benchmark tested clinical reasoning from text vignettes; it did not assess multimodal or visual models. Extending the same evidence--choice--confidence framework to vision-language systems would be valuable, particularly for tasks involving spatial reasoning, radiological interpretation, or other forms of visual evidence integration. One possibility is that multimodal models may show different or stronger dissociations between evidence, decision, and confidence than language-only models, including spatial biases, weaker evidence integration, or poorer confidence calibration. Such extensions would help determine whether the partial metacognitive sensitivity observed here is specific to linguistic clinical reasoning or generalizes across modalities.

Future work should extend the current AT-NCD versus DRCI contrast while expanding the dataset, especially around the ambiguity boundary where the model showed its clearest breakdown. This should include more vignette seeds, more graded evidence levels near the diagnostic boundary, and a larger set of conflicting and missing-information cases. A larger dataset would allow more stable estimation of calibration, AUROC2, and exploratory SDT-based measures such as meta-d$^\prime$/d$^\prime$.

Future work should also extend the benchmark across more models and diagnostic contrasts. Testing models from different families and capability levels would make it possible to ask whether confidence--correctness discrimination improves with model scale, architecture, or training regime, rather than assuming that stronger first-order performance implies better uncertainty monitoring. 

More broadly, this framework could support the development of a leaderboard for artificial metacognitive efficiency in medical LLMs and beyond. Such a leaderboard should not rely on a single score. It should report complementary quantities, including diagnostic accuracy, calibration error, Brier score, information-sufficiency performance, confidence modulation by evidence level, and confidence--correctness discrimination. AUROC2 is useful because it provides an accessible summary of how well confidence separates correct from incorrect responses on the same controlled benchmark. However, AUROC2 becomes less informative when model accuracy approaches ceiling, because it requires both correct and incorrect trials to estimate confidence--correctness discrimination. As seen with \texttt{gpt-5.5}, a model may show meaningful confidence modulation across evidence levels while producing too few errors for AUROC2 to be defined or stable.

For this reason, future leaderboards should treat AUROC2 as one component of artificial metacognitive evaluation rather than as a complete measure. As models become more accurate, closer inspection of how confidence tracks graded evidence, conflicting information, missing information, and underdetermined cases will become increasingly important. More elaborate measures such as $\mathrm{meta}$-$d^\prime$/$d^\prime$ may also be useful when datasets are large and difficult enough to support stable estimation, but they should be reported alongside transparent behavioral measures rather than replacing them.

\section{Conclusion}

This study introduced a controlled clinical-evidence benchmark for evaluating diagnostic choice, confidence calibration, and information-sufficiency judgments in a medical LLM. Diagnostic choices were strongly evidence-sensitive, and confidence was not globally flat or arbitrary. Confidence increased with evidence strength, decreased when key information was missing, and remained higher on correct than on incorrect trials even after accounting for evidence quality and prompt format. These findings indicate that explicit confidence in a medical LLM can carry meaningful second-order information. Medical-LLM evaluation should not stop at overall diagnostic accuracy or broad benchmark scores. Controlled evidence manipulations make it possible to separate first-order diagnostic sensitivity from second-order confidence behavior and to identify local regions in which otherwise competent models become biased or miscalibrated. The exploratory model comparison further suggests that newer or nominally stronger models should not be assumed to have better metacognitive sensitivity. Confidence quality must be measured directly, rather than inferred from model generation, scale, or benchmark accuracy alone. In that sense, the present work provides a proof of concept for psychophysics-inspired evaluation of medical-LLM uncertainty and for future benchmarks of artificial metacognitive efficiency.

\section*{Data and code availability}

All synthetic vignettes, prompt templates, trial manifests, model outputs, analysis notebooks, and figure-generation code are available at:

\noindent\url{https://github.com/anazz-dev/medical-llm-metacognition-benchmark}.

\appendix
\section{Supplementary analyses}

Supplementary analyses included a penalized logistic model of AT-NCD choice as a robustness check on evidence-sensitive first-order behavior and exploratory SDT-based efficiency estimates, including $d^\prime$, meta-$d^\prime$, and meta-$d^\prime$/$d^\prime$. These analyses supported the main interpretation but were not treated as primary because the pilot sample was small, several cells were at or near ceiling, and confidence values occupied a restricted range.

\section{Supplementary analyses}

Supplementary analyses included a penalized logistic model of AT-NCD choice as a robustness check on evidence-sensitive first-order behavior and exploratory SDT-based efficiency estimates, including $d^\prime$, $\mathrm{meta}$-$d^\prime$, and $\mathrm{meta}$-$d^\prime$/$d^\prime$. These analyses supported the main interpretation but were not treated as primary because the pilot sample was small, several cells were at or near ceiling, and confidence values occupied a restricted range.

\begin{table}[H]
\centering
\caption{Supplementary penalized logistic model predicting AT-NCD choice.}
\label{tab:supp_logistic_choice}
\small
\begin{tabular}{lrr}
\toprule
Predictor & Coefficient & Odds ratio \\
\midrule
Conflicting information & 0.332 & 1.393 \\
Missing information & -0.442 & 0.643 \\
DRCI-first prompt (P02) & -0.730 & 0.482 \\
Named-options prompt (P03) & 0.527 & 1.693 \\
Evidence score & 0.706 & 2.025 \\
Evidence $\times$ conflicting & -0.102 & 0.903 \\
Evidence $\times$ missing & -0.028 & 0.973 \\
\bottomrule
\end{tabular}
\end{table}

\noindent
The penalized model confirmed that diagnostic choice was strongly evidence-sensitive. Each one-point increase in AT-NCD evidence approximately doubled the odds of choosing AT-NCD. The DRCI-first prompt shifted responses away from AT-NCD, and missing-information cases also reduced the odds of an AT-NCD response. Interaction terms were small, suggesting that the main effect was a shift in decision boundary rather than a loss of evidence sensitivity.

\begin{table}[H]
\centering
\caption{Supplementary first-order and metacognitive sensitivity estimates.}
\label{tab:supp_metacognitive_efficiency}
\small
\begin{tabular}{lr}
\toprule
Metric & Estimate \\
\midrule
Forced-choice trials & 108 \\
Type-1 hit rate, Hautus-corrected & 0.864 \\
Type-1 false-alarm rate, Hautus-corrected & 0.009 \\
Type-1 sensitivity, $d^\prime$ & 3.46 \\
Type-1 criterion, $c$ & 0.63 \\
AUROC2 & 0.876 \\
Exploratory $\mathrm{meta}$-$d^\prime$ & 1.07 \\
Exploratory M-ratio, $\mathrm{meta}$-$d^\prime$/$d^\prime$ & 0.31 \\
\bottomrule
\end{tabular}
\end{table}

\noindent
The SDT estimates indicate strong first-order diagnostic discrimination, with high $d^\prime$ and a conservative AT-NCD response criterion. AUROC2 indicated substantial confidence--correctness discrimination. However, exploratory $\mathrm{meta}$-$d^\prime$ was substantially lower than $d^\prime$, yielding an M-ratio of approximately 0.31. This suggests that the model's confidence signal captured only part of the information available to the primary diagnostic decision process.

\begin{table}[H]
\centering
\caption{Supplementary model comparison using AUROC2.}
\label{tab:supp_model_auroc2}
\small
\begin{tabular}{lrr}
\toprule
Model & Forced-choice trials & AUROC2 \\
\midrule
\texttt{gpt-4.1-nano} & 108 & 0.873 \\
\texttt{gpt-4.1-mini} & 108 & 0.786 \\
\texttt{gpt-5-nano} & 108 & 0.644 \\
\bottomrule
\end{tabular}
\end{table}

\noindent
The exploratory model comparison suggests that confidence--correctness discrimination did not improve monotonically with nominal model capability or newer model family. In this benchmark, \texttt{gpt-4.1-nano} showed the highest AUROC2, followed by \texttt{gpt-4.1-mini}, while \texttt{gpt-5-nano} showed weaker confidence discrimination. This comparison should be interpreted cautiously because the benchmark was small and focused on one diagnostic contrast, but it supports the broader point that confidence quality should be measured directly rather than inferred from model identity or benchmark accuracy.

\begin{figure}[H]
    \centering
    \includegraphics[width=0.6\linewidth]{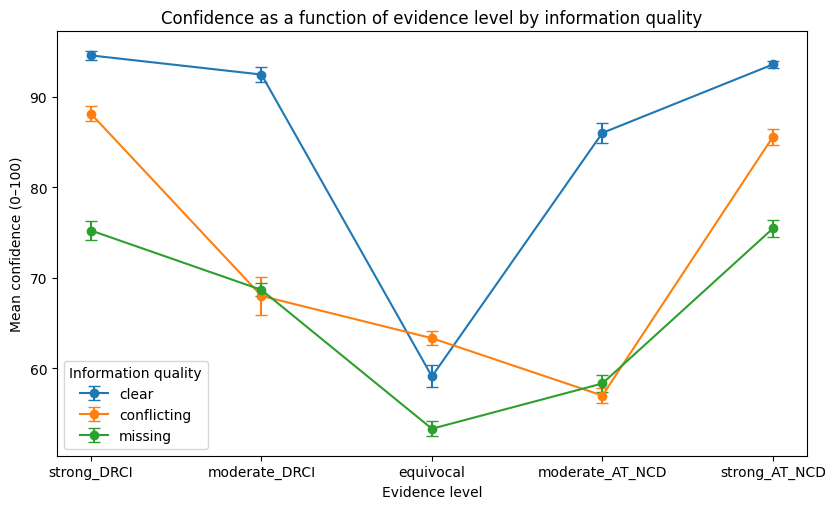}
    \caption{\textbf{Confidence modulation in \texttt{gpt-5.5}.}
    Although AUROC2 could not be estimated for \texttt{gpt-5.5} because the model made no forced-choice errors, confidence still varied across evidence levels and information-quality conditions. This shows that confidence modulation by evidence and confidence--correctness discrimination are related but distinct quantities.}
    \label{fig:supp_gpt55_confidence_evidence}
\end{figure}

\bibliographystyle{plainnat}
\bibliography{refs}

\end{document}